# Cross-strait Variations on Two Near-synonymous Loanwords *xie2shang1* and *tan2pan4*: A Corpus-based Comparative Study

**Yueyue Huang**[1,2]

[1] Guangzhou Xinhua University, 19 Huamei Road, Tianhe District, Guangzhou, Guangdong, China;

[2] The Hong Kong Polytechnic University, Kowloon, Hong Kong SAR, China

`hailey.huang@connect.polyu.hk`

**Chu-Ren Huang**

Department of Chinese and Bilingual Studies, The Hong Kong Polytechnic University, Kowloon, Hong Kong SAR, China

`churen.huang@polyu.edu.hk`

**Abstract**

This study attempts to investigate cross-strait variations on two typical synonymous loanwords in Chinese, i.e. 协商(*xie2shang1*) and 谈判 (*tan2pan4*), drawn on MARVS theory[1]. Through a comparative analysis, the study found some distributional, eventual, and contextual similarities and differences across Taiwan and Mainland Mandarin. Compared with the underused *tan2pan4*, *xie2shang1* is significantly overused in Taiwan Mandarin and vice versa in Mainland Mandarin. Additionally, though both words can refer to an inchoative process in Mainland and Taiwan Mandarin, the starting point for *xie2shang1* in Mainland Mandarin is somewhat blurring compared with the usage in Taiwan Mandarin. Further on, in Taiwan Mandarin, *tan2pan4* can be used in economic and diplomatic contexts, while *xie2shang1* is used almost exclusively in political contexts. In Mainland Mandarin, however, the two words can be used in a hybrid manner within political contexts; moreover, *tan2pan4* is prominently used in diplomatic contexts with less reference to economic activities, while *xie2sahng1* can be found in both political and legal contexts, emphasizing a role of mediation.

## 1 Introduction

Research into near-synonyms in Mandarin Chinese, particularly verbal ones, has attracted scholarly attention in recent years. Exploration of semantic differences of near-synonyms can contribute to our knowledge of the Chinese language. As "syntactic behaviours of verbs are semantically determined" (Chief et al., 2000, p. 57), a comparison of the syntactic information of synonymous verbs thus can effectively reveal semantic differences between the verbs (Chief et al., 2000).

Nonetheless, the syntactic information is sometimes intricated to catch and can be confusing to scholars of interest for lacking a clear representation of semantic clues hidden in syntactic information. In response to it, the Module-Attribute Representation of Verbal Semantics (MARVS) theory was then proposed to construct Chinese verbal semantics better. The theory was based on the premise that lexical semantic representation is the grammaticalization of conceptual information, i.e., they can be linked to grammatical structure with conceptual motivation and be attested by representational clues (Chung & Ahrens, 2008; Huang et al., 2000; Huang & Hsieh, 2015). Representational clues include collocation, argument section constraints, distributional patterns along with other elements that can be attested by corpus evidence (Huang & Hsieh, 2015).

MARVS theory denotes that verbal semantics can be differentiated based on eventive information, which is comprised of event modules and role modules, both bearing its internal attributes (see Figure 1). There are five 'atomic event

---

[1] The hereafter number "1, 2, 3, 4" followed after Pinyin corresponds to the four Chinese tone mark [ ¯ ´ ˇ ` ].

structures' in the event modules, including (Huang et al., 2000, p.26):

(1)**.** Boundary: it can be identified with a temporal point, and that must be regarded as a whole.

(2) **/** Punctuality: a single occurrence of an activity that cannot be measured by duration.

(3) **/////** Process: an activity that has a time course.

(4) **_____** State: a homogeneous module in which the concept of temporal duration is irrelevant.

(5) **^^^^^** Stage: a module which consists of iterative sub-events.

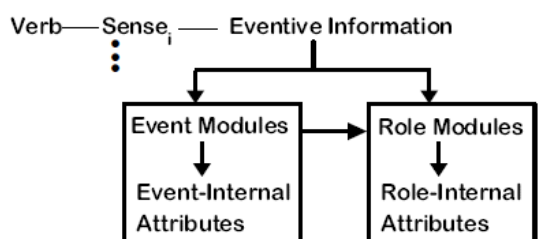

Figure 1. Theoretical framework of MARVS (Huang et al., 2000)

The role modules include the focused roles (participants) of an event and the role-internal attributes, the latter being "the semantic properties of the participants, such as [sentience], [volition], and so forth" (Huang & Hsieh, 2015)

Under the framework of MARVS, a considerable number of studies have been carried out on Chinese verbal near-synonyms (Ahrens et al., 2003; Chung & Ahrens, 2008; Tao, 2021; Wang & Huang, 2018). Further studies on MARVS expanded our understanding of aspects related to Chinese verbal near-synonyms, including their spatial-temporal attributes (Liang & Huang, 2021) and mental states (Tao, 2021),

Additionally, as cross-strait variation has also been observed by scholars (Hong & Huang, 2008; Hung et al., 2007), a more specific focus on issues related to such variations using MARVS has been made to enrich our views on the diversity within the Chinese language. A few issues have been touched upon, such as power relation (Wang & Huang, 2018) and viewpoint foci differentiation (Wang & Huang, 2021). Nonetheless, fine-grained semantic relations on cross-strait variations are still in need of further investigation.

This study then attempts to contribute to our understanding of cross-strait variations on synonymous loanwords in Chinese, taking two typical loanwords as an entry point, i.e. 协商(*xie2shang1*) and 谈判 (*tan2pan4*). MARVS theory was adopted as the analytical framework to construct a fuller picture of the semantic variations of the two words.

## 2 The present study

**2.1 Loan words in Chinese**

Chinese orthographical or syntactic system has not been static. It has undergone generations of evolutions with both in-group and out-group momentum. One typical out-group linguistic influence comes from cross-cultural contacts, hence ubiquitous loanwords observed in Chinese language systems. Loanwords could come from English (Kim, 2018), Japanese (Shi, 2020), and many other languages (e.g. Russian) that might have come into contact with China throughout history.

In particular, a large number of political or social terms were introduced in the Chinese language from the late 1800s to the 1900s, a time of social turmoil (Gunn, 1991; Masini, 1993). The social terms 谈判 (*tan2pan4*) and 协商 (*xie2shang1*), listed in official Chinse loanwords dictionaries (Shi, 2019, p.1108+1253), were introduced into modern Chinese roughly at such time. The former was considered to be translated into Chinese through relay translation (English to Japanese to Chinese) in the period of 1840~1920s (from the Opium War of 1840 to before the Anti-Japanese War) (Shi, 2019, p.1108; Chen, 2014); while the later regarded as a merged, interchangeable term with both Japanese and Chinese word 协议 (*xie2yi1*, agreement), which has been influenced from Japanese in the late 1800s as well (Shi, 2019, p.1253), and eventually formed what we observed in modern usage as it might later have mixed influence from a Latin word, *deliberationem* (Wang & Zhang, 2010).

To confirm what was found in the literature, a balanced one billion-bytes CCL corpus [2] and Google N-gram Viewer[3] were chosen to examine the traces of their usage in history. The CCL corpus (Center for Chinese Linguistics PKU) consists of 581,794,456 modern Chinese characters and 201,668,719 ancient Chinese characters (Zhan et al., 2019). Google N-gram is an online search engine that charts the frequencies of search in printed books between 1500 and 2019 in text corpora collected by Google in eight languages, including Chinese (Michel et al., 2011).

---

[2] http://ccl.pku.edu.cn:8080/ccl_corpus/

[3] https://books.google.com/ngrams

The two words were first searched in the ancient Chinese subcorpus of CCL. The term *xie2shang1* generated 63 hits, with the earliest usage traced back to the late Qing Dynasty. And the search for *tan2pan4* retrieved 76 hits with the earliest usage found in the period of the Republic of China. Selected examples in CCL are listed as follows:

Example 1: 丁亥，命班第赴金川军营**协商**军务。(二十五史\清史稿)

Pinyin: *ding1hai4, ming4 ban1di4 ban1di4 fu4 jin1chuan1 jun1ying2 **xie2shang1** jun1wu4.*

Translation: In the year of Dinghai, Bandi was ordered to go to the Jinchuan military camp to **consult on** military affairs.

Example 2: 要求贵国即予同意，……速开正式**谈判**……（民国\小说\民国演义）

Pinyin: *yao1qiu2 gui4guo2 ji2yu3 tong2yi4, ... su4 kai1 zheng4shi4 **tan2pan4**...*

Translation: We hope that you can agree to open a formal **negotiation** soon.

modern Chinese around the second half of the 19th century.

## 2.2 Research Questions

As the two words (i.e., *xie2shang1* and *tan2pan4*) were established as loan words, the present study was then valid to proceed. The study attempts to analyze cross-strait variations of these two synonymous loan words using MARVS theory. More precisely, it attempts to answer the following research questions:

RQ1: What are the distributional differences of *xie2shang1* and *tan2pan4* between Mainland Mandarin and Taiwan Mandarin?

RQ2: What are the event representations of *xie2shang1* and *tan2pan4* in Mainland Mandarin and Taiwan Mandarin?

RQ3: What are the role representations of *xie2shang1* and *tan2pan4* used in Mainland Mandarin and Taiwan Mandarin?

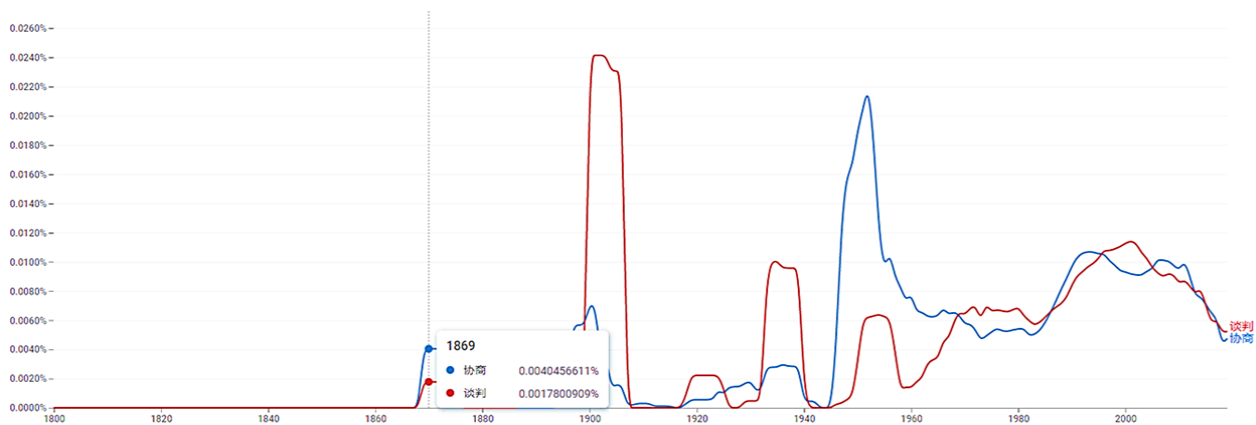

Figure 2 Google N-gram results for *xie2shang1* and *tai2pan4*

To double-check the result, the authors also generated Google N-grams of the two words, as shown in Figure 2. Both words appeared roughly at the same time in the 1870s, which fall into the generalized findings in the previous literature. Subsequent searches through Google Books provided examples in Japanese texts, but, through screening, the earlier attested examples for the two words in Chinese were found in 清議報 (*Ching i Po*, Qingyi Newspaper), published by the Royalists led by Liang Qichao in Japan in 1900. Cross-reference of all sources, including literature, CCL corpus, Google Books, and Google N-gram results, suggested that the two words came into

## 2.3 Research Method

The Gigaword Corpus (CWS)[4] and its two subcorpora were chosen for this study — Gigaword_XIN (XIN) and Gigaword_CNA (CNA) via Chinese Word Sketch (Hong et al., 2006; Ma et al., 2006). The former (XIN) was compiled by news texts from Xinhua News Agency of Beijing (382,881,000 tokens), and the latter (CNA) by news from the Central News Agency of Taiwan (735,499,000 tokens) (Huang & Wang, 2020).

In line with the MARVS-based lexical semantics methodology proposed in the studies of (Chung & Ahrens, 2008; Huang et al., 2000), the

---

[4] https://wordsketch.ling.sinica.edu.tw/

present study will follow the research process to answer the above questions as stated:

First, to establish the near-synonymous relationship of the two words by analyzing their senses based on the meanings in the Chinese WordNet and examining the examples in the main corpus.

Second, to examine the distributional patterns of the two words in Mainland and Taiwan Mandarin.

Third, to analyze their collocations (e.g., modifier/modified, propositional phrases) to construct their event representations.

Forth, to analyze the agent-goal/subject-object relationship and the role internal attributes of the two words to construct their role representations.

## 3 Findings and Discussions

### 3.1 Establishing near-synonyms

It is somewhat tricky to establish the near-synonymous semantic relations between *xie2shang1* and *tan2pan4*. In Chinese WordNet (CWN)[5], compiled by the Institute of Linguistics, Academia Sinica, *xie2shang1* means "reaching a consensus through discussion among disagreeing parties" (意见不同的几方一起讨论以取得各方都能接受的结论; *yi2jian4 bu4tong2 de ji1fang1 yi4qi3 tao3lun4 yi3 qu3de2 ge4fang1 dou1 neng2 jie1shou4 de jie2lun4*). It can serve both as an intransitive VERB and a NOUN. The definition provided by CWN suggests that the word is an action involving more than two parties, with the potential of a mediator, and it aims for a win-win, optimizing outcome for all.

In the case of *tan2pan4*, CWN does not manage to compose its corresponding meaning description, yet is able to indicate this word is a combination of two morphemes *tan2,* meaning "to exchange information in a verbal manner" (用言语交换讯息; *yong4 yu3yan2 jiao1huan4 xun4xi1*) and *pan4,* meaning "to make a conclusion about the subsequent events based on certain criteria" (根据特定标准对后述事件做出结论;*gen1jun4 te4ding4 biao1zhun3 dui4 hou4shu4 shi4jian4 zuo4chu1 jie2lun4*). In this sense, the word can consequently mean "to make a conclusion about certain events through conversation or discussion". It thus refers to an action that might typically involve two parties, optimizing the outcome for either party (a zero-sum game).

Examination for concordances in CWS confirmed certain interchangeability of the two words. For example:

1) ……将继续和市府及业者**协商**，使运价更合理。

Pinyin: *...jiang1 ji4xu4 he2 shi4fu3 ji2 ye4zhe3 **xie2shang1**, shi3 yun4jia4 geng4 he2li3*.

Translation: ...will continue **negotiating** with the city government and the industry to make the tariff more reasonable.

2) 不断完善共产党领导的多党合作和政治**协商**制度…

Pinyin: *bu4duan4 wan2shan4 gong4chan3dang3 ling3dao3 de duo1dang3 he2zuo4 he2 zheng4zhi4 **xie2shang1** zhi4du4…*

Translation: …continue to improve the system of multi-party cooperation and political **consultation** led by the Communist Party…

3) 如果美国不与伊拉克**谈判**，伊拉克就不会退出科威特。

Pinyin*: ru2guo3 mei3guo2 bu4 yu3 yi1la1ke4 **tan2pan4**, yi1la1ke4 jiu4 bu4hui4 tui4chu1 ke1wei1te4*.

Translation: Iraq will not withdraw from Kuwait if the US does not **negotiate** with Iraq.

Since the near-synonymous relationship of the two words is established, it is then to analyze the cross-strait variations in terms of their distributions, event modules and role modules.

### 3.2 Distributional Variations

To examine their cross-strait variations, we searched two node words in both subcorpora CNA and XIN, and the result can be seen in Table 1. As the two subcorpora contain a disproportionate number of corpus data, a log-likelihood formula[6] was run to compare the frequency distribution across the two sub-corpora.

Table 1 Distributional variations of *xie2shang1* and *tan2pan4* in CNA and XIN

|  | Freq. in CNA | Freq. in XIN | Log-Likelihood | sig. |
|---|---|---|---|---|
| 谈判 | 111,619 | 67,301 | 894.52 | 0.000 *** - |
|  | **Freq. in CWS** | | | 180,550 |
| 协商 | 91,998 | 20,215 | 14604.35 | 0.000 *** + |
|  | **Freq. in CWS** | | | 112,649 |

---
[5] https://cwn.ling.sinica.edu.tw/
[6] https://ucrel.lancs.ac.uk/llwizard.html

The results suggest that, compared with the significantly underused tan2pan4, xie2shang1 is significantly overused in Taiwan and vice versa in Mainland China.

Additionally, CWS PoS results indicate that *tan2pan4* can serve as both activity transitive verbs (VC2) and as a common noun (Na), while *xie2shang1* can only be activity verbs with a sentential object (VE2). Both actions indicate the dual participation of agent as subject and goal as the object. The grammatical category of *xie2shang1* found in Gigaword Corpus seems to show a discrepancy with what was concluded in CWN, issued by a prestigious linguistic institute in Taiwan, which suggests it can be both a verb and a noun. The concordance search in XIN was then run to confirm whether there is a valid discrepancy in the grammatical categories of xie2shang1 between Taiwan and Mainland Mandarin. It was found that the word, though sometimes tagged as VE2, could still present deverbal features in context, such as in No.0005 in Figure 3. It implies that the word might be in a blurring grammatical position between verbal and deverbal elements in Mainland Mandarin.

| | |
|---|---|
| 0005 | ，以及與河北省有關方面的 協商/VE2，公司已經投資買下１１·６ |
| 0071 | 每年根據不同情況，經過 協商/VE2 解決。目前，南朝鮮每年 |
| 0012 | 之上，國際事務應由各國 協商/VE2 解決，而不應由一兩個大國 |
| 0012 | 並希望兩國繼續擴大政治 協商/VE2 和加強經濟合作。錢外長 |
| 0024 | 建議在平壤或漢城舉行統一 協商/VE2 會議 新華社 平壤 １月 ８日電 |

Figure 3 Concordances of *xie2shang1*(協商/协商) in XIN

One possible explanation for this might be that the overuse of *xie2shang1* in Taiwan Mandarin broadens the scope of its grammatical categories, while underusage of *xie2shang1* in Mainland Mandarin may still experience an ongoing and changing process for the word's grammatical attributes. This further points to a subtle linguistic change that might occur with a disproportionate usage of the same language in different geographical locations.

### 3.3 Event Representation

WordSketch for each word was run in both XIN and CNA to examine their semantic variations in collocation. As each word yielded different occurrences in two subcorpora, the nominalized frequency was provided to establish a comparable baseline and multiplied by 10,000 for the purpose of clearer presentation.

All collocated results were examined to construct the event modules of the words, including the returned lists of their possessors, possessions, modifiers, the modified and propositional phrases. Typical collocations indicating one of the 'atomic event structures' were extracted, and the top four frequent ones were shown in Table 2 and Table 3. Table 2 included one more collocated word, "中" (*zhong1;* middle), as it typically occurs as part of the syntactic structure "正在……中……" (*zheng4zai1…zhong1*; being in the middle of…).

Table 2 Collocation of *tan2pan4* across CNA and XIN

| 谈判 | collocation | F. | NF* | T-score | MI |
|---|---|---|---|---|---|
| **CNA =111,619** | 迄 | 12 | **1.08** | 3.46 | 14.15 |
| | 今 | 12 | **1.08** | 3.46 | 12.54 |
| | 开始 | 245 | 21.95 | 15.65 | 11.24 |
| | 正在 | 153 | 13.71 | 12.37 | 12.46 |
| | 中 | 26 | 2.33 | 5.01 | 5.77 |
| | 进展 | 137 | 12.27 | 11.70 | 14.08 |
| **XIN =67,301** | 迄 | 12 | **1.78** | 3.46 | 13.91 |
| | 今 | 12 | **1.78** | 3.46 | 13.18 |
| | 开始 | 385 | 57.21 | 19.61 | 10.46 |
| | 正在 | 74 | 11.00 | 8.60 | 11.13 |
| | 中 | 30 | 4.46 | 5.43 | 6.83 |
| | 过程 | 212 | 31.50 | 14.56 | 12.43 |

*NF: Nominalized frequency=Frequency/10,000*

Table 2 shows that both in Taiwan and Mainland Mandarin, the event structures of *tan2pan4* point to both boundary and process. For its boundary structure, the word *tan2pan4* has a clear starting point, as *kai2shi3* (开始; start to) can be used with *tan2pan4* (MI=11.24/10.46); however, its ending point is not very clear as the only one collocated propositional phrase *qi4jin1* (迄今; so far) only occur roughly ONCE in 10,000 times either in Taiwan or Mainland Mandarin. Furthermore, the phrase 'so far' tends to point to a middle point during such an event process. Its process indicator is rather prominent as it can both be collocated with process indicators, such as 正在谈判中 (*zheng4zai4 tan2pan4 zhong1*; is under negotiation…), 谈判进展 (*tan2pan4 jin4zhan3*; negotiation progress) or 谈判过程 (*tan2pan4 guo4cheng2*; the process of negotiation). Thus *tan2pan4* can be considered an inchoative process in Taiwan and Mainland mandarin ( • /////).

The collocations with the word *xie2shang1* similarly indicate that this word can refer to a process event with a starting point, yet no ending point (see Table 3). Typical process indicators other than 正在(*zheng4zai4*), which was mentioned above, also include 继续 (*ji4xu4*; continue to...), 持续 (*chi2xu4*, continue to…), etc.. Additionally, though both Taiwan and Mainland usage of *xie2shang1* have the indicator (*kai2shi3*) for its starting point of a boundary, comparing the nominalized frequency of *kai2shi3* suggests that the starting point in its Taiwan usage is prominently clearer than that in its Mainland Mandarin (NF in CNA=41.85 > NF in XIN=8.41). So even though the word can be an inchoative process in Taiwan Mandarin ( • /////) and Mainland Mandarin, the starting boundary in Mainland Mandarin ( • /////) is somewhat blurring, comparatively speaking.

Table 3 Collocation of *xie2shang1* across CNA and XIN

| 协商 | collocation | F. | NF* | T-score | MI |
|---|---|---|---|---|---|
| **CNA =91,998** | 继续 | 911 | 99.02 | 30.17 | 11.15 |
| | 持续 | 68 | 7.39 | 8.24 | 11.87 |
| | 正在 | 75 | 8.15 | 8.66 | 12.46 |
| | 开始 | 385 | **41.85** | 19.61 | 10.46 |
| **XIN =20,215** | 继续 | 55 | 27.21 | 7.41 | 10.89 |
| | 坚持 | 31 | 15.34 | 5.57 | 11.70 |
| | 正在 | 57 | 28.2 | 7.55 | 11.13 |
| | 开始 | 17 | **8.41** | 4.12 | 10.46 |

*NF: Nominalized frequency=Frequency/10,000*

It is also worth mentioning that both words in XIN and CNA have disposal inherent event attributes, as both can be collocated with *ba,* for example:

(1) ……一直反对把贸易谈判跟其他政治问题扯在一起。
Pinyin: *yi1zhi2 fan3dui4 ba3 mao4yi4 tan2pan4 gen1 qita1 zheng4zhi4 wen4ti2 che3 zai4 yi1qi3...*
Translation: …has always been opposed to tying trade talks with other political issues.
(2) 把总预算案协商，是为政治角力工具。
Pinyin: *ba3 zong3 yu4suan4 an4 xie2shang1, shi4wei2 zheng4zhi4 jue2li4 gong1ju4.*
Translation: … to take the general budget consultation as a tool of political wrestling.

Findings in this session point to little distinctions of the words across Mainland and Taiwan Mandarin. It is understandable as the two words were introduced to the Chinese language only roughly over a century ago, and they have not undergone many changes throughout time. Nonetheless, their role representations might well exhibit quite diverged contextual variations as the words experienced social turmoil since their introduction. The following session will compare their role representations in XIN and CNA.

### 3.4 Role Representation

Common patterns (Figure 4 and 5) and Only patterns (Table 4 and 5) for agent-goal/object-subject relationship were then generated to examine the contextual role representations of the two words in Taiwan and Mainland Mandarin.

The common patterns (Figure 4 and 5) in two different regions reveal that the two node words seem to share more semantic common ground in their Mainland usage as both of their collocated subjects and objects in XIN have more commonality than those in the CNA corpus. In XIN, there are many collocations with political implications (e.g., 政党, *zheng4dang3*, political parties; 领导人 *ling3dao3 ren2*, political leaders) found in common patterns; while in CNA, only words bearing no social or political implication (i.e., 我方, *wo3fang1*, our party; 双方, *shuang1fang1*, two parties) were found to share common collocational patterns between the two words. It implies that the two synonyms are interchangeably used in political contexts in Mainland usage.

Additionally, however, in Taiwan Mandarin, the two words are treated with different contextual implications as *xie2shang1* are almost exclusively used in political contexts (e.g. 政党, *zheng4dang3*, political parties; 朝野, *chao2ye3*, government), while *tan2pan4* can be used in a broader range of activities, including political, economic or social settings (e.g. 政治性, *zheng4zhi4xing4*, political attributes; 资方, *zi1fang1*, investor; 财产权, *cai2chan3quan2*, property rights). Such difference might be the result of several rounds of language education reforms in mainland China in the past few decades, leading to a more hybrid usage of near-synonyms in Mainland Mandarin (Mills, 1956; Sheridan, 1981).

Figure 4 Common patterns of *xie2shang1* and *tan2pan4* in CNA

Figure 5 Common patterns of *xie2shang1* and *tan2pan4* in XIN

A further scrutinization of the object-subject relationship for only patterns of these two words across CNA and XIN echoed the above findings and revealed more clues for contextual usage (see Table 4 and 5).

In CNA, *tan2pan4* can be found in both economic and diplomatic contexts (e.g. 埃雷卡特 Saeb Erakat, a Pakistan diplomat). Additionally, further examination of their concordance lines confirms such findings, such as 新回合谈判 (*xin1hui2he2 tan2pan4*, new round of negotiation). However, *xie2shang1* is almost exclusively related to political contexts (see Table 4).

Table 4 Only patterns of *tan2pan4* and *xie2shang1* in CNA

| tan2pan4 | | | |
|---|---|---|---|
| Subject | Freq. | NF. | MI |
| 贸易 | 1383 | 123.90 | 47.4 |
| 首席 | 439 | 39.33 | 53 |
| 航权 | 306 | 27.41 | 57.1 |
| 新回合 | 201 | 18.01 | 60.6 |
| 世界贸易组织 | 115 | 10.30 | 29.6 |
| Object | Freq. | NF. | MI |
| 代表团 | 500 | 44.80 | 37 |
| 龙永图* | 200 | 17.92 | 53.8 |
| 对手 | 167 | 14.96 | 29.2 |
| 埃雷卡特# | 132 | 11.83 | 55 |
| 程序性 | 88 | 7.88 | 42.1 |
| | Total freq. | | =111,619 |
| xie2shang1 | | | |
| Subject | Freq. | NF. | MI |
| 人民 | 582 | 63.26 | 32.7 |
| 单位 | 372 | 40.44 | 23.8 |
| 党政 | 265 | 28.80 | 42.1 |
| 三党 | 83 | 9.02 | 32.7 |
| 预备性 | 23 | 2.50 | 26.2 |
| Object | Freq. | NF. | MI |
| 制度 | 120 | 13.04 | 19.7 |
| 版本 | 102 | 11.09 | 34.8 |
| 总预算案 | 85 | 9.24 | 32 |
| 会报 | 61 | 6.63 | 23.5 |
| 修正案 | 35 | 3.80 | 16.9 |
| | Total freq. | | =91,998 |

(*a Chinese economist; # a diplomatic representative for Pakistan)

In XIN, *tan2pan4* is primarily found in diplomatic contexts with less apparent reference to economic activities, while *xie2sahng1* can be found in both political and legal contexts (see Table 5). Additionally, objects for *xie2shang1* in Mainland usage can emphasize the exchanges of discussion being conducted with a possible mediator during such a process (e.g. 座谈会, *zuo4tan2hui4*, discussion panel; 委员, *wei3yuan2*, committee member).

Table 5 Only patterns of *tan2pan4* and *xie2shang1* in XIN

| tan2pan4 | | | |
|---|---|---|---|
| **Subject** | **Freq.** | **NF.** | **MI** |
| 回合 | 475 | 70.58 | 58.8 |
| 首席 | 454 | 67.46 | 56.3 |
| 地位 | 453 | 67.31 | 40.4 |
| 入盟 | 320 | 47.55 | 61.5 |
| 阶段 | 227 | 33.73 | 30.3 |
| **Object** | **Freq.** | **NF.** | **MI** |
| 进展 | 286 | 42.50 | 34.1 |
| 立场 | 206 | 30.61 | 31 |
| 埃雷卡特# | 223 | 33.13 | 59.6 |
| 龙永图* | 133 | 19.76 | 52.1 |
| 僵局 | 110 | 16.34 | 37.2 |
| | Total freq. | | =67,301 |
| xie2shang1 | | | |
| **Subject** | **Freq.** | **NF.** | **MI** |
| 人民 | 197 | 97.45 | 18.3 |
| 部门 | 134 | 66.29 | 18.7 |
| 单位 | 39 | 19.29 | 11 |
| 当事人 | 37 | 18.30 | 27.7 |
| 事务性 | 17 | 8.41 | 25.4 |
| **Object** | **Freq.** | **NF.** | **MI** |
| 委员 | 49 | 24.24 | 19.6 |
| 对话 | 23 | 11.38 | 17.2 |
| 职能 | 22 | 10.88 | 18 |
| 座谈 | 19 | 9.40 | 21.7 |
| 座谈会 | 19 | 9.40 | 21.7 |
| | Total freq. | | =20,215 |

Being loan words, such diverse contextual usage settings of *xie2shang1* and *tan2pan4* used in Mainland and Taiwan Mandarin might point to a sociological *status quo*, requiring further studies in disciplines of historical linguistics. It is also worth noting that no specific markers bearing role-internal attributes were found in their collocation across XIN and CNA.

## 4 Conclusion

*Xie2shang1* and *tan2pan4*, as loan words, share synonymous common grounds for "reaching to some conclusions through discussions" with parties involved in such a process. Nonetheless, a comparative study of the two near-synonyms based on CNA and XIN of the CWS corpus reveals some distributional, eventual, and contextual similarities as well as differences across Taiwan and Mainland Mandarin.

Their distributional patterns suggested that, compared with the significantly underused *tan2pan4*, *xie2shang1* is significantly overused in Taiwan Mandarin, and vice versa in Mainland Mandarin. On their event representations, distinctions were not found between Mainland and Taiwan Mandarin as both words can refer to an inchoative process, though the starting point for *xie2shang1* is rather blurring compared with that in Taiwan Mandarin. It might be in relation to their relatively 'young' status within modern Chinese vocabulary.

Nonetheless, an examination of the subject-object/ agent-goal relationship of the two words revealed more details on their different contextual usages in Mainland and Taiwan Mandarin. The two words share more semantic common ground, or, more precisely, in political contexts of Mainland usage, *tai2pan4* and *xie2shang1* are used in a hybrid manner. Additionally, in Mainland Mandarin, *tan2pan4* is found more prominently used in diplomatic contexts with less apparent reference to economic activities, while *xie2sahng1* can be found in both political and legal contexts, emphasizing a possible mediator. In contrast, the object-subject relationship in Taiwan Mandarin suggests the two words are used in quite a different context. In Taiwan Mandarin, *tan2pan4* can be used in economic and diplomatic contexts, while *xie2shang1* is used exclusively in political contexts.

The role of contextual and distributional differences might point to other historical or sociological factors that may play a certain role in decerning the variational linguistic changes throughout time and history. Nonetheless, this does not fall into the role of the present study. Further investigations on a diachronic basis might contribute to our understanding of how and what loan words might evolve or change throughout historical, social, or political events.

**Acknowledgements**

   The work has been funded by the 2019 Research Project for Young Scholars of Guangzhou Xinhua University (2019KYQN11; On the Collaborative Effects of Working Memory and Background Information on L2 Reading), and the Higher Educational Reform Project of Guangzhou Xinhua University (2019T001; Construction of Translation Teaching Team in School of Foreign Languages)